\documentclass[twocolumn,10pt]{wlscirep}
\usepackage[T1]{fontenc}
\usepackage{array}
\usepackage{makecell}
\usepackage{diagbox}
\usepackage{tabularx}
\usepackage{colortbl}
\usepackage[ruled,vlined]{algorithm2e}
\usepackage[skip=3pt,font=small]{subcaption}
\usepackage{mathtools}
\usepackage[capitalize,nameinlink]{cleveref}
\usepackage{xspace}
\usepackage{cuted}

\usepackage{amsmath, amsfonts, multirow}
\usepackage{amsthm} % 提供 proof 环境
\usepackage{tcolorbox} % 加载 tcolorbox 包
\tcbuselibrary{theorems} % 使用 theorems 库以启用 \tcbhighmath 命令
\usepackage{csquotes}
\makeatletter
\newcommand{\thickhline}{%
    \noalign {\ifnum 0=`}\fi \hrule height 1pt
    \futurelet \reserved@a \@xhline
}
\makeatother
\definecolor{DarkBlue}{RGB}{64,101,149}
\definecolor{azure}{rgb}{0.0, 0.5, 1.0}
\definecolor{gray}{rgb}{0.3, 0.3, 0.3}
\definecolor{DarkGreen}{RGB}{42,110,63}
\definecolor{myred}{RGB}{255,0,0} % 红色
\definecolor{mygray}{gray}{.9}
\usepackage{pifont}

\crefname{algorithm}{Alg.}{Algs.}
\Crefname{algocf}{Alg.}{Algs.}
\crefname{section}{Sec.}{Secs.}
\Crefname{section}{Section}{Sections}
\crefname{table}{Tab.}{Tabs.}
\Crefname{table}{Table}{Tables}
\crefname{figure}{Fig.}{Figs.}
\Crefname{figure}{Figure}{Figures}

\title{Co-policy: Responsive Human-Robot Co-Creation for Musical Performances}
\author[1,+]{Xuetao Li}
\author[5,+]{Wenke Huang}
\author[1]{Mang Ye}
\author[2]{Zijian Liu}
\author[3]{Jinhua Xie}
\author[1,*]{Jifeng Xuan}
\author[1,4,*]{Miao Li}

\affil[1]{School of Computer Science, Wuhan University, Wuhan, China}
\affil[2]{School of Automation, Wuhan University of Technology, Wuhan, China}
\affil[3]{School of Geodesy and Geomatics, Wuhan University, Wuhan, China}
\affil[4]{School of Robotics, Wuhan University, Wuhan, China}
\affil[5]{College of Computing and Data Science, Nanyang Technological University, Singapore}
\affil[+]{Joint First Authors}
\affil[*]{Correspondence: jxuan@whu.edu.cn, miao.li@whu.edu.cn.

 Project webpage: \textcolor{magenta}{\url{https://xtli12.github.io/Co-policy/}}}

\begin{abstract}
Art has long stood as a pivotal expression of human creativity. Embodied artificial intelligence offers a route for generative models to participate in that creativity through physical action rather than disembodied digital content. In robotic music co-creation, it is challenging to connect semantic musical understanding with real-time and physically executable performance. We present \textbf{Co-policy}, a framework for human-robot musical co-creation that separates semantic intent grounding, constrained musical variation, and visuomotor execution. To ground musical semantics, Co-policy uses pre-inference semantic anchors and a fine-tuned Qwen-vl planner (F-Qwen) to transform speech, live musical seeds, and visual observations into structured co-creation plans. To support low-latency execution, Co-policy introduces a Gaussian-Mixture Visuomotor Policy (GMP), implemented as a conditional mixture-density policy that maps target notes and visual context to multimodal robot actions in a single forward pass. Unlike robotic playback systems that merely reproduce user-specified notes, Co-policy generates complementary musical responses under both musical and physical constraints. Real-robot chime experiments, ablations, and expert evaluation show improved intent alignment, execution accuracy, and response frequency over diffusion-policy and ablated baselines, supporting physically grounded action generation as a key requirement for embodied human-AI co-creation.
\end{abstract}

\begin{document}

\flushbottom
\maketitle

\thispagestyle{empty}
\clearpage

% \newpage
% \clearpage
% \input{./checklist}

% \begin{links}
% \link{Code}{https://anonymous.4open.science/r/CoPo-DCF0}
% \end{links}

\section{Introduction}
\label{sec:intro}
The intersection of artificial intelligence (AI) and the arts has witnessed significant advancements in recent years, particularly in domains such as music, visual arts, and creative writing \cite{sturm2024ai,liu2024sora,wang2024weaver}. Most generative AI systems, however, remain disembodied: they produce text, images, audio, or symbolic scores, but do not have to act through a physical body, negotiate instrument constraints, or respond at the time scale of human interaction. Embodied musical co-creation provides a compact but demanding testbed for this next step. A physically situated agent must understand a human's abstract intent, generate a musically meaningful response, and realize that response through timed contact with an instrument.
Despite this opportunity, robotic music systems are often closer to playback devices than co-creative agents: a human specifies notes, and the robot executes them. This leaves two scientific questions unresolved. \textbf{I)} \textit{How can a robot turn an incomplete human seed into a complementary musical response?} \textbf{II)} \textit{How can that response be executed physically, accurately, and with interaction-level latency?} Diffusion policies are highly expressive and have recently achieved very strong performance in many robotic manipulation tasks~\cite{chi2023diffusion}. However, their iterative denoising process can introduce latency that is undesirable for tightly timed human-robot musical interaction \cite{dong2024diffuserlite,chen2024simple}. While accelerated diffusion schemes (e.g., consistency models or fast samplers) mitigate this, they often trade off mode coverage for speed. Meanwhile, standard imitation models trained with near-duplicate views and repeated poses often collapse toward mean actions, losing the alternative striking modes needed for expressive performance.
% Additionally, the physical embodiment of robots in artistic activities demands dexterous and adaptable hardware to mimic the nuanced movements inherent in human artistry \cite{ashton2024people,schaldenbrand2024cofrida}. 

To tackle these challenges, we present \textbf{Co-policy}, a lightweight embodied AI agent that fuses F-Qwen, a fine-tuned Qwen-vl vision-language model (VLM), a constrained musical variation planner, and a low-latency Gaussian-Mixture Visuomotor Policy (GMP). Co-policy converts the disembodied foundation model Qwen-vl into a responsive embodied agent: the VLM grounds intent, the planner creates motif-, harmony-, novelty-, and playability-constrained musical content, and GMP executes it through timed strikes. This modular vision-language-action (VLA) design avoids the deployment cost of monolithic end-to-end VLA models while retaining a physically grounded action loop.

\textbf{Rethinking ``co-creation''.} A central premise of our work is that genuine co-creation is \emph{not} mechanical replay of a human-specified score. We define embodied musical co-creation as a closed-loop interaction in which: (i) the human provides an incomplete creative seed, such as a motif, style, rhythm, or verbal intent; (ii) the AI agent interprets this seed under musical and physical constraints; (iii) the robot contributes a complementary response rather than directly copying the input; and (iv) the response is physically instantiated through real-time embodied action, with timing, strike dynamics, and spatial adaptation constrained by the robot and instrument. This definition yields four measurable dimensions: intent alignment, creative contribution, embodied feasibility, and interaction responsiveness.

To address \textbf{I)}, we construct a semantic anchor bank for Qwen-vl~\cite{yang2024qwen2} consisting of paired musical seeds, score images, style descriptors, technical annotations, robot-playability tags, and target structured plans. During inference, retrieved anchors are inserted into a fixed prompt template, and the VLM outputs a JSON-formatted co-creation plan rather than unconstrained prose. The downstream planner then generates a robot response that preserves the human motif while introducing harmonic, rhythmic, or accompaniment variation. This design balances the emergent semantic reasoning of the VLM with structural guardrails for physical execution.

To address \textbf{II)}, we introduce GMP as a single-pass conditional mixture-density policy. Given the egocentric visual observation and the target note plan, our GMP directly predicts mixture weights, means, and covariances over full robot action vectors (or short-horizon trajectory segments). The mixture components represent latent action modes--alternative striking postures, approach directions, and contact timings--rather than individual joints. This preserves multiple feasible physical responses while avoiding the sequential denoising steps of diffusion policies.

% A key innovation of our network lies in the fusion of four distinct modalities: speech, music stream inputs, visual imagery, and robotic actions. By incorporating these diverse data streams, our GMP achieves a robust and flexible understanding of the creative context, allowing the music robot to generate coherent and contextually appropriate musical performances. This multimodal fusion not only enhances the robot's ability to respond to real-time human inputs but also supports the generation of complex musical sequences that align with the creative intentions of human collaborators.

% Furthermore, we have engineered a flexible dexterous hand that significantly improves the robot's ability to perform musical tasks with human-like precision. Our hand design employs flexible materials and a rope-driven mechanism to emulate the muscle and skin tissues of the human hand, resulting in more natural and expressive movements. This innovation ensures that the robot's interactions with musical instruments, such as striking chimes, produce sounds that closely resemble those generated by human hands, thereby enhancing the authenticity and aesthetic quality of the performances.

We evaluate Co-policy on real-robot chime performance and use ManiSkill2 as a secondary generalization sanity check. The real-robot study focuses on whether the agent aligns with human intent, contributes new musical material, executes feasible actions, and responds at interactive speed. In summary, our contributions are threefold: 
\begin{itemize}[leftmargin=*, itemsep=2pt, topsep=3pt, parsep=0pt, partopsep=0pt]
\item \textbf{\textit{A formal embodied co-creation presentation.}} We formulate musical co-creation as a closed-loop interaction problem in which a robot turns an incomplete human creative seed into a complementary musical response that is semantically aligned and physically executable, distinguishing it from robotic playback or disembodied generation.
\item \textbf{\textit{A modular embodied agent for constrained musical variation.}} Co-policy combines semantic intent grounding, constrained musical variation, and visuomotor execution. The VLM outputs structured plans from speech, live notes, and visual context, while the planning layer constrains generated responses by motif consistency, harmonic validity, novelty, and robot playability.
\item \textbf{\textit{A low-latency Gaussian-Mixture Visuomotor Policy.}} We propose GMP, implemented as a conditional mixture-density policy, to predict multimodal action distributions over full robot action vectors in a single forward pass. This avoids iterative denoising while preserving alternative feasible action modes. We validate the system through real-robot experiments, ablation studies, latency analysis, and blinded expert evaluation.
\end{itemize}

\begin{figure*}[t]
    \centering
    \includegraphics[width=\textwidth]{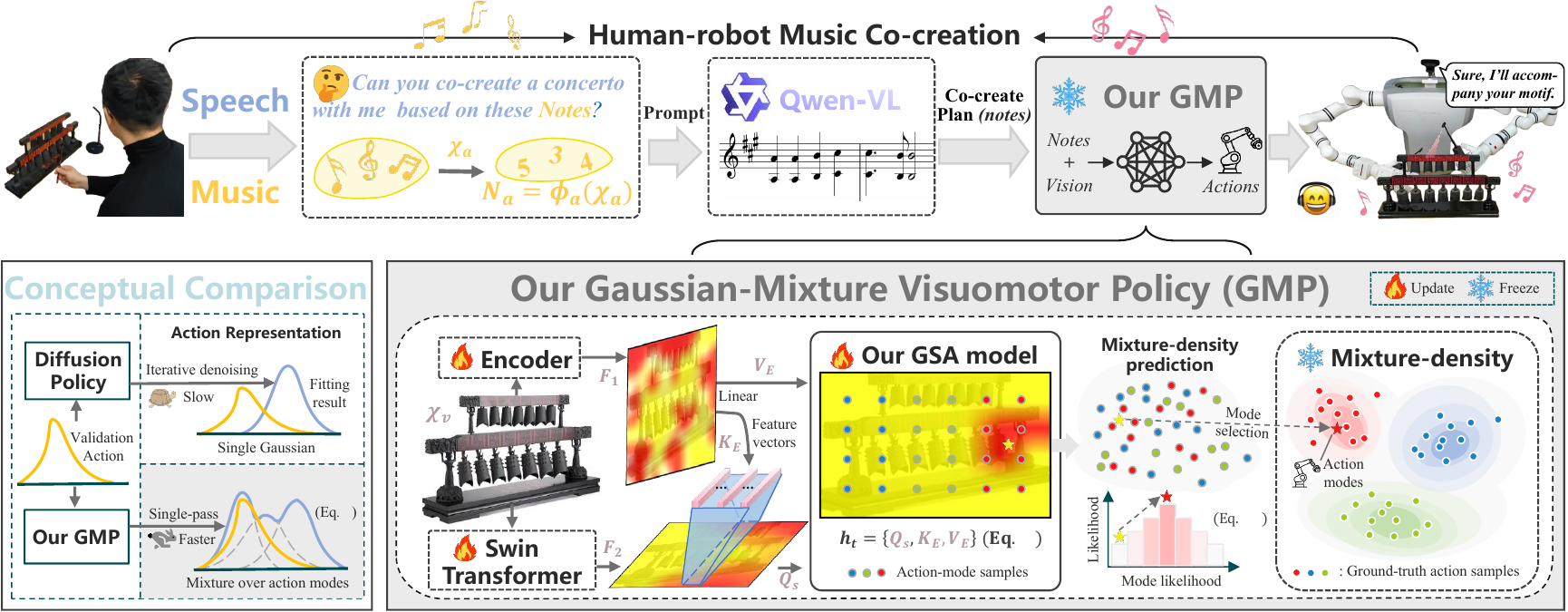}
            % \put(-187.5,83.4){\makebox(0,0)[l]{\fontsize{8}{9}\selectfont\textbf{\ref{subsec:GMP}}}}
            \put(-175.5,23.5){\makebox(0,0)[l]{\fontsize{6}{7}\selectfont \ref{eq:GSNET}}}
           \put(-105.3,30){\makebox(0,0)[l]{\fontsize{5}{6}\selectfont\ref{eq:regression}}}
             \put(-392.5,31.5){\makebox(0,0)[l]{\fontsize{5}{6}\selectfont\ref{eq:fitting}}}
        \vspace{-0.2cm}
         \caption{\textbf{Overview of Co-policy.}
Our framework processes multimodal inputs (human speech, live musical seeds \(\chi_a\), and RGB observations) to support embodied human-robot co-creation. The VLM grounds the human's incomplete creative seed into a structured intent plan; the constrained musical variation module generates a complementary robot response; and the \textbf{G}aussian-\textbf{M}ixture \textbf{V}isuomotor \textbf{P}olicy (\textbf{GMP}) maps the response and visual context to executable robot actions in a single forward pass. Compared to diffusion-policy control, the conditional mixture-density policy preserves alternative action modes without iterative denoising; representative video demonstrations are provided on the accompanying project webpage.}
\label{fig:pipeline}
\vspace{-0.4cm}
\end{figure*}

% Our method not only enhances the capabilities of AI-driven music robots but also contributes to the broader conversation on AI and creativity. It underscores the potential for advanced AI systems to collaborate with human artistic creators, fostering new forms of creative expression.

\subsection{Background and Related Work}
\label{sec:related}

\subsubsection{Human-robot Co-creation Framework}
Human-robot co-creation (HRC) merges robotics, AI, and the creative arts to enable rich, collaborative interaction. Early Human-Robot Interaction (HRI) research focused on predefined tasks \cite{dautenhahn2005social}, evolving towards dynamic systems that perceive and respond to human behavior in real-time \cite{wamba2023bibliometric,hashmi2023artificial}. Creative robotics now spans music, visual art, and beyond, while collaborative-creativity frameworks employ deep and reinforcement learning to generate content autonomously yet coherently with human partners \cite{karimi2020creative,hsieh2022conversational}.
Multi-modal interaction enables robots to interpret various forms of human expression. Affective computing has been integrated to recognize and respond to human emotions \cite{de2023co,de2024co}. Models like InstructBLIP \cite{instructblip} and LLaVA \cite{liu2024visual,liu2023improved} have improved image-text integration. However, their use in robotics applications faces considerable challenges due to real-world variability, platform heterogeneity, and the necessity for reliable action control \cite{shridhar2023perceiver,team2024octo}. We use Qwen-vl \cite{yang2024qwen2} as a semantic grounding module. Robot-specific states and constraints, such as reachable bells, timing limits, and playability rules, are exposed through structured prompts and downstream planning rather than assumed to be directly modeled by the VLM.
 
\subsubsection{Action Generation Framework}
Learning-based motion planning is central to advanced robotic manipulation \cite{Khandate-RSS-23,Li-RSS-23,chi2023diffusion}. \textit{Reinforcement learning} achieves strong performance but depends on carefully crafted motion primitives and reward functions \cite{kim2023pre,xu2021efficient}. \textit{Imitation learning} reduces reward design by copying expert demonstrations, yet its generalization is bounded by demonstration quality and coverage \cite{Haldar-RSS-23,decisiontransformer,shafiullah2022behavior}. Diffusion models have recently been adopted as generative action planners that produce trajectories through iterative score-based optimization \cite{huang2023diffusion,li2024source,xian2023chaineddiffuser}. Although expressive, their repeated denoising steps can limit inference rate in tightly timed human-robot interaction \cite{dong2024diffuserlite,evans2024fast}. To address this bottleneck, we introduce \textbf{GMP}: a plug-and-play, goal-conditioned controller that combines a Guided Self-Attention encoder with a conditional mixture-density action head. GMP operates on RGB inputs and target note plans, producing multimodal joint commands in a single forward pass without denoising.

\section{Results}
\label{sec:results}
We evaluate Co-policy from three complementary perspectives: whether the agent generates semantically aligned and musically complementary responses, whether the generated note plans can be executed accurately by a real robot, and whether the visuomotor policy retains interactive response speed under matched baselines. The results show that semantic anchoring improves co-creation quality, while the Gaussian-mixture action head improves real-robot execution without diffusion-style repeated denoising.

\begin{figure}[t]
% \Centering
  \includegraphics[width=\columnwidth]{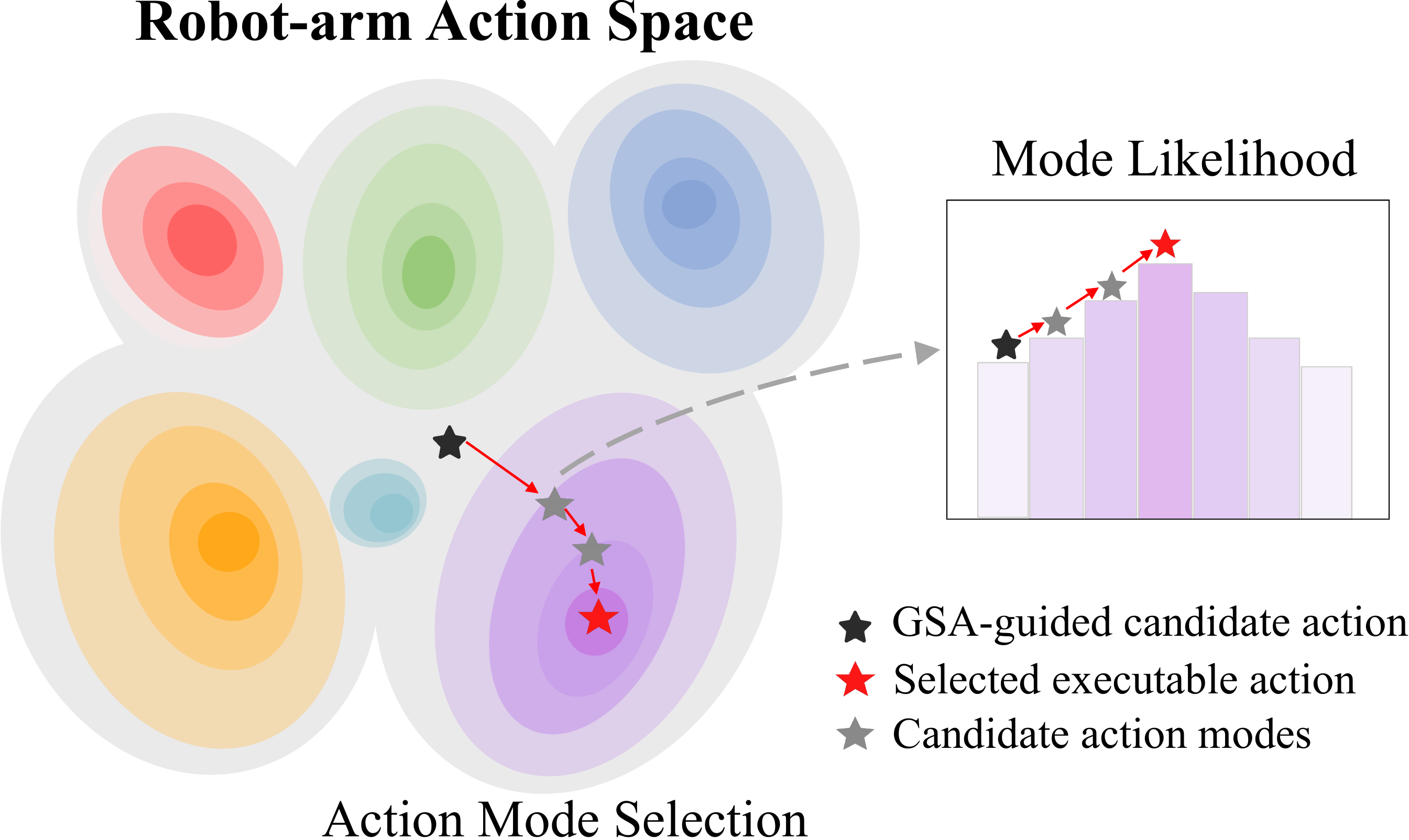}
  \vspace{-0.6cm}
  % \captionsetup{justification=justified, singlelinecheck=false}
  \caption{\textbf{Illustration of latent action modes in GMP.} Each Gaussian component represents one feasible action mode over the full \(H\)-step six-DoF action segment, such as a top-down strike, side swing, or gentle tap. Components do not correspond to individual joints. At inference time, GMP predicts mixture parameters in a single forward pass and selects or aggregates the executable action without EM fitting, gradient update, or iterative refinement.}
\label{gmm}
\vspace{-0.4cm}
\end{figure}

\begin{figure*}[t]
	\centering
	\includegraphics[width=1.96 \columnwidth]{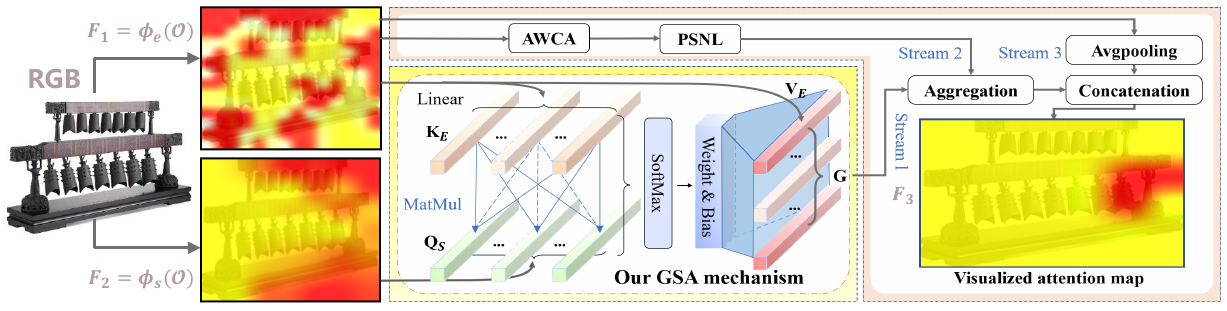}
        \vspace{-0.3cm}
	\caption{\textbf{Detailed structure of our GSA model.} Unlike vanilla self-attention, our GSA couples global interaction modeling with locally salient features across distinct maps: local features (red in \(F_1\), processed by encoder \(\phi_e\)) denote key regions in the observation, while global features (red in \(F_2\), via Swin Transformer \(\phi_s\)) capture long-range regional connections. Integrating AWCA (Adaptive Weighting Channel Attention) and PSNL (Patch-level Second-order Non-Loc) module for channel-wise and global feature alignment, the GSA enables accurate localization of the target chime bell (red in \(F_3\)).}
	\label{fig:GSA}
    \vspace{-0.5cm}
\end{figure*}

\subsection{Problem Formulation and Overview}
Real-time human–robot musical co-creation demands that a robot comprehend \textbf{\emph{what}} to play (semantic and musical reasoning over speech and live notes), \textbf{\emph{what to add}} (a complementary response rather than a copy), and \textbf{\emph{how}} to play it (geometric and contact reasoning over visual cues). At each instant~$t$, the robot consumes a multimodal observation \mbox{$(\mathcal{I}_t, N^u_{t}, \mathcal{O}_t)$}: the spoken instruction $\mathcal{I}_t$, the human-provided seed notes $N^u_t$ extracted from the audio stream~$\chi_{a,t}$, and the egocentric RGB frame $\mathcal{O}_t\!\in\!\mathbb{R}^{H\times W\times3}$. The agent outputs a robot response $N^r_t$ and a physically executable action segment $\tau_t=[a_t,\ldots,a_{t+H}]\in\mathbb{R}^{H\times6}$.

\paragraph{What constitutes embodied musical co-creation?}
We define co-creation as a four-part closed loop: (1) the human provides an incomplete creative seed $N^u_t$ or verbal intent $\mathcal{I}_t$; (2) the agent interprets the seed under semantic, musical, and physical constraints; (3) the robot generates a complementary response $N^r_t$ with measurable novelty relative to $N^u_t$; and (4) the response is physically realized through real-time embodied action $\tau_t$. This formulation separates co-creation from two simpler settings: robotic playback, where $N^r_t=N^u_t$, and disembodied music generation, where the output is not constrained by a physical robot or instrument.
\subsubsection{Why Guided Self-Attention}
\textbf{Motivation.} 
In robotic tasks, establishing an end-to-end mapping between the robotic end-effectors and targets is crucial for efficient action generation. Prevalent DP-based iterative methods suffer from slow inference, while traditional end-to-end transformers prioritize global dependencies but neglect local pixel-wise features that are essential for fine-grained motion planning. To address this, we propose an end-to-end visuomotor model, Guided Self-Attention (GSA), which first \textit{decouples} global and local features from the same observation $\mathcal{O}_t$, and then applies guided attention to \textit{couple} global reasoning with the relevant local features, mitigating over-reliance on long-range dependencies while establishing the end-to-end mapping mechanism.

% In robotic tasks, the precise capture of spatial relationships in visual inputs is essential for associating end-effector of robot with action targets. Traditional self-attention mechanisms tend to emphasize global dependencies while neglecting local pixel-wise features that are critical for fine-grained motion planning. This imbalance hampers the effectiveness of vision-based motion prediction, as existing transformer models fail to reconcile global context with the preservation of local details, thereby unable to establish the spatial relationship between end-effector of robot and action targets precisely. To address this challenge, we propose the Guided Self-Attention (GSA) model, which first \textit{decoupled} global and local features from the same observation $\mathcal{O}_t$, and then applies guided attention to \textit{couple} global reasoning with relevant local features, mitigating the over-reliance on long-range connections.

\paragraph{Design rationale of guided self-attention.}
We do not claim a formal guarantee that attention must perfectly localize physical contacts. Instead, GSA is designed to bias global contextual reasoning toward locally salient contact regions. The Swin stream provides a full-scene representation of the instrument layout, while the DenseNet stream preserves contact-scale local descriptors. By taking queries from the global stream and keys/values from the local stream,
\begin{equation}
G(F_s, F_e) = \mathrm{softmax}\!\left( \frac{Q_S K_E^{\!\top}}{\sqrt{d_k}} + B \right) V_E,
\end{equation}
GSA lets the policy ask \enquote{which part of the instrument is the target?} using global context, but answer through local evidence that corresponds to reachable bells and contact regions. We empirically verify this design through attention visualization, ablation of the global/local streams, and failure analysis rather than relying on an over-strong theorem.

\subsubsection{Why GMP uses a mixture-density action head}
\textbf{Motivation.} In robotic manipulation datasets, near-duplicate viewpoints and similar target articulations create clustering artifacts that drive networks toward mean actions and erode motion diversity. We counteract this collapse by modeling the conditional distribution over the full robot action vector or short-horizon trajectory segment, rather than predicting a single deterministic mean. Given visual context $\mathcal{O}_t$, a robot note plan $N^r_t$, and physical constraints $c_t$, the policy predicts:
\begin{equation}
    p_\theta(\tau_t \mid \mathcal{O}_t,N^r_t,c_t)
    =
    \sum_{k=1}^{K}\pi_k
    \mathcal{N}\!\left(\tau_t;\mu_k,\Sigma_k\right),
\end{equation}
where $\tau_t\in\mathbb{R}^{H\times6}$ can be reduced to a single-step action when $H{=}1$. The mixture components do \textbf{not} correspond to individual joints. Each component represents a latent action mode, such as an alternative striking posture, approach direction, or contact timing pattern that can realize the same note under different visual configurations. In the current implementation, we use $K{=}6$ as a practical capacity hyperparameter rather than as a proxy for the robot's six degrees of freedom; systematic selection of $K$ is left to future work. This formulation preserves multiple feasible embodied responses while remaining a single-forward-pass policy, offering an analytical alternative to accelerated diffusion models that often sacrifice mode coverage for inference speed.

% \paragraph{Evaluation Metrics.}
% For each trial we record  

% \begin{itemize}
% \item[]\hspace{-1em}%
% $P$ — \emph{pitch precision}, i.e.\ the ratio of correctly struck notes;
% \item[]\hspace{-1em}%
% $F=1/\tau$ Hz — \emph{response frequency}, inverse of average inference time~$\tau$;
% \item[]\hspace{-1em}%
% $M$ — \emph{melodic quality}, obtained by averaging  
%       (i)~consonance ratio with Western intervals and  
%       (ii)~a 5-point expert mean-opinion score.
% \end{itemize}

% The overall co-creation score is
% \begin{equation}
% \mathcal{S} \;=\; P + \beta F + \gamma M,
% \quad
% \beta = 0.05,\;
% \gamma = 0.25.
% \end{equation}\small

\subsection{Co-policy Framework}
We present \textbf{Co-policy}, a framework for real-time human-robot musical co-creation that integrates multimodal understanding (music-language-visual) with precise motion generation. Our approach leverages egocentric RGB images and real-time music streams to enable accurate 6D robotic control for expressive instrument performance. The whole pseudocode for our pipeline is presented in Algorithm~\ref{alg:GMP}. The following sections formalize Co-policy and describe its training procedure.

\begin{algorithm}[!t]
\caption{Co-policy training and inference}
\label{alg:GMP}
\SetNoFillComment
\SetArgSty{textnormal}
\small
\KwIn{Human instruction $\mathcal{I}_t$, human seed notes $N_t^u$, RGB observation $\mathcal{O}_t$, semantic anchor bank $\mathcal{S}$, demonstrations $\mathcal{D}$}
\KwOut{Robot note response $N_t^r$ and executable action segment $\tau_t^*$}

\BlankLine
\tcp{Training}
Collect $\mathcal{D}=\{(\mathcal{O}_i,N_i^r,c_i,\tau_i^{\mathrm{gt}})\}_{i=1}^{M}$, where $\tau_i^{\mathrm{gt}}\in\mathbb{R}^{H\times6}$\;
\For{$i=1,\ldots,M$}{
    $h_i\leftarrow \mathrm{GSA}(\mathcal{O}_i)$\;
    $\{\pi_k,\mu_k,\Sigma_k\}_{k=1}^{K}\leftarrow f_\theta(h_i,N_i^r,c_i)$\;
    $\mathcal{L}_{\mathrm{MDN}}\leftarrow -\log\sum_{k=1}^{K}\pi_k\mathcal{N}(\tau_i^{\mathrm{gt}};\mu_k,\Sigma_k)$\;
    $\mathcal{L}\leftarrow \mathcal{L}_{\mathrm{MDN}}+\lambda\left\|\sum_{k=1}^{K}\pi_k\mu_k-\tau_i^{\mathrm{gt}}\right\|_2^2$\;
    Update $\theta$ by gradient descent on $\mathcal{L}$\;
}

\BlankLine
\tcp{Inference}
\For{each interaction step $t$}{
    $\mathcal{Z}_t\leftarrow \mathrm{Parse}_{\mathrm{VLM}}(\mathcal{I}_t,N_t^u,\mathcal{O}_t,\mathcal{S})$\;
    $c_t\leftarrow \mathrm{EstimateConstraints}(\mathcal{O}_t)$\;
    % \tcp*{reachable bells, occlusions, tempo, workspace}
    $N_t^r\leftarrow f_{\mathrm{plan}}(N_t^u,\mathcal{Z}_t,c_t)$\;
    % \tcp*{complementary response, not replay}
    $h_t\leftarrow \mathrm{GSA}(\mathcal{O}_t)$\;
    $\{\pi_k,\mu_k,\Sigma_k\}_{k=1}^{K}\leftarrow f_\theta(h_t,N_t^r,c_t)$\;
    % \tcp*{single forward pass}
    $\tau_t^*\leftarrow \mu_{\arg\max_k \pi_k}$ or $\tau_t^*\leftarrow\sum_{k=1}^{K}\pi_k\mu_k$\;
    Execute $\tau_t^*$ on the robot\;
}
\end{algorithm}

\subsubsection{Embodied semantic grounding and planning}
\label{subsec:vlm}
At the highest level of the system, the VLM is not treated as the sole source of creativity. Instead, it is one component in an embodied planning pipeline that separates semantic intent grounding, constrained musical variation, and physical execution. We employ Qwen-vl~\cite{bai2023qwen} with a semantic anchor bank whose entries contain a score image, symbolic notes, style descriptor, meter/tempo annotation, robot-playability tags, and an expected structured plan. At inference, retrieved anchors enter a fixed prompt.

\textbf{1) Semantic intent parsing.}
The VLM converts speech, live seed notes, and visual context into a structured representation:
\[
\mathcal{Z}_t =
\mathrm{Parse}_{\mathrm{VLM}}(\mathcal{I}_t,N^u_t,\mathcal{O}_t,\mathcal{S}),
\]
where $\mathcal{Z}_t$ contains fields such as style, tempo, meter, key, human seed, robot role, reachable bells, and maximum response latency. We require the VLM to output JSON rather than free-form text, for example:
\begin{quote}\scriptsize\ttfamily
\{ intent: "create", style: "energetic", tempo: 120,\\
human\_seed: [3,5,6,5], robot\_role: "accompaniment",\\
available\_notes: [1,2,3,5,6], max\_latency\_ms: 500 \}.
\end{quote}

\textbf{2) Constrained musical variation.}
The robot response is then generated as a complementary note plan
\[
N^r_t = f_{\mathrm{plan}}(N^u_t,\zeta_t,c_t),
\]
where $\zeta_t$ is the style/intent latent extracted from $\mathcal{Z}_t$, and $c_t$ represents physical constraints estimated from the observation, such as reachable bells and tempo limits. We constrain $N^r_t$ by four criteria: motif consistency (retaining recognizable structure from the human seed), harmonic validity (avoiding tonal conflict), novelty (preventing direct copying), and embodied playability (ensuring the plan can be executed within workspace and latency limits).

\textbf{3) Embodied execution.}
Finally, the generated note plan is mapped to robot motion by the visuomotor policy:
\[
\tau_t \sim \pi_\theta(\tau\mid \mathcal{O}_t,N^r_t,c_t),
\]
where $\tau_t$ is a joint action segment. This separation is important: semantic anchoring improves intent grounding, while the creative planner and policy determine what the robot contributes and how it physically realizes that contribution.

The formal representation of the planning module is refined as follows:
\begin{equation}
\label{eq:vlm}
\mathcal{P}_t = plan\left(\mathcal{I}_t, N^u_t, \mathcal{O}_t \middle| \mathcal{C}, \mathcal{S}\right),
\end{equation}
where $\mathcal{S} = \{s_1, s_2, \ldots, s_n\}$ denotes the semantic anchor bank and \(\mathcal{C}\) contains the fixed schema and playability rules. This framework makes the planning interface reproducible and inspectable through the implementation protocol described in the Methods.  

\paragraph{Constrained creative contribution beyond replay.}
Crucially, the human collaborator never specifies the full output sequence. As distilled in the prompt template of Figure~\ref{fig:prompt_main}, the human provides only a semantic anchor (a target mood) together with a short seed motif and the live tempo, while the VLM is asked to \emph{complete} rather than \emph{copy}. Conditioned additionally on the visually observed instrument state (which bells are reachable and currently free of residual vibration), the model fills in harmonically consistent responses, adds an accompaniment line, and emits per-note dynamics that track the human's beat. Figure~\ref{fig:prompt_main} contrasts two regimes for the identical seed motif \texttt{C4 E4 G4}: a non-anchored baseline that mechanically echoes the three input notes, versus our anchored agent that, given the anchor \enquote{joyful, allegro}, extends the motif into a \texttt{C--G--Am--Em} response with a syncopated accompaniment and stronger down-beat strikes while avoiding the occluded F bell. This constrained variation, which is absent from the seed and filtered by physical playability, is what we operationalize as the robot's creative contribution.

\begin{figure}[t]
\centering
\begin{tcolorbox}[colback=mygray!25,colframe=DarkBlue,boxrule=0.4pt,
  left=3pt,right=3pt,top=2pt,bottom=2pt,fonttitle=\bfseries\small,
  title=Co-creation prompt template (distilled)]
{\scriptsize\ttfamily
[System] You are a musical co-creator, not a transcriber.\\{}
[Anchor $\mathcal{S}$] mood=\{joyful\}; style=\{allegro, 4/4\}\\{}
[Motif $N_a$] seed\_notes = C4 E4 G4 ; tempo = 120 BPM\\{}
[Vision $\mathcal{O}$] free\_bells = \{C,D,E,G,A\}; occluded = \{F\}\\{}
[Task] Complete \& accompany the motif; output notes,\\{}
\phantom{[Task] }right/left-hand assignment, and strike dynamics.\par}
\vspace{2pt}\hrule\vspace{2pt}
{\scriptsize
\textbf{Baseline (replay):} \texttt{C4 E4 G4} (echoes input).\\{}
\textbf{Ours (co-creative):} response \texttt{C--G--Am--Em} with
syncopated accompaniment and accented down-beats, avoiding the occluded F bell.}
\end{tcolorbox}
\vspace{-4pt}
\caption{\textbf{Distilled co-creation prompt and constrained creative response.} The human supplies only a mood anchor and a short motif; conditioned on the visually observed instrument state, the anchored VLM \emph{completes}, not \emph{replays}.}
\label{fig:prompt_main}
\vspace{-0.5cm}
\end{figure}

% The integration of semantic anchoring and dynamic prompt composition represents a key innovation over standard fine-tuning approaches, as it allows the model to generalize musical knowledge across unseen scores while maintaining real-time responsiveness critical for interactive musical co-creation scenarios.

% This process is illustrated in the model framework Figure~\ref{fig:pipeline}. 

% \subsection{Speech-to-Text Conversion Method}
% \label{subsec:speech}
% To facilitate seamless human-robot interaction, we developed a speech-to-text pipeline integrating real-time audio capture, accurate speech recognition, and dynamic robotic control. The system uses the sound device library to capture high-fidelity audio at 16,000 Hz and processes it with the Vosk toolkit for transcription. Post-processing corrects common errors, enhancing accuracy. For feedback, the robot provides verbal feedback using Microsoft SAPI.

\subsubsection{Gaussian-Mixture Visuomotor Policy}
\label{subsec:GMP}
% \begin{figure}[t]
% % \Centering
%   \includegraphics[width=\columnwidth]{IAWCA.png}
  
%   % \captionsetup{justification=justified, singlelinecheck=false}
%   \caption{Structure of IAWCA, which can fully exploit channel-wise feature responses by integrating correlations between channels of feature maps.}
% \label{fig:IAWCA}
% \vspace{-0.4cm}
% \end{figure}

Our Gaussian-Mixture Visuomotor Policy (GMP) bridges visual perception, generated note plans, and robotic action. GMP is implemented as a \textbf{single-pass conditional mixture-density policy}: given the egocentric RGB observation $\mathcal{O}_t$, the robot-generated note plan $N^r_t$, and physical constraints $c_t$, the policy predicts a multimodal distribution over full robot actions (or short-horizon trajectory segments). This lets the controller preserve multiple feasible striking modes while meeting the low-latency requirements of musical interaction.

\paragraph{Guided Self-Attention model.}
Robotic tasks demand precise spatial reasoning from visual inputs, requiring association of scene regions with end-effector positions for motion planning. However, conventional transformer self-attention \cite{vaswani2017attention} overemphasizes global dependencies while overlooking local pixel-wise features critical for fine-grained motion. This creates a critical yet challenging global-local imbalance in motion prediction. To resolve this, we propose the GSA model, a dual-stream architecture with two key innovations: \textbf{1}) \textit{Decoupled Feature Extraction} and \textbf{2}) \textit{Cross-Stream Coupling Guided Attention}.

First, decoupled feature extraction explicitly separates global and local feature learning: a Swin Transformer stream (\(\phi_s\)) captures long-range spatial dependencies via shifted windows \cite{liu2022swin} (e.g., scene-level structure), while a parallel DenseNet stream extracts pixel-wise independent local features (e.g., subtle motion cues). This design retains both global context and fine-grained details, resolving the trade-off in conventional transformers.

Second, cross-stream coupling guided attention ensures global interactions are grounded in local relevance. Unlike vanilla self-attention (where queries, keys, and values derive from the same feature map) GSA pairs query vectors (\(Q_S\)) from the global Swin stream with key/value vectors (\(K_E, V_E\)) from the local DenseNet stream. This coupling directs global reasoning toward task-relevant local features, mitigating over-reliance on long-range connections. The formulation of the guided self-attention is as follows:
\begin{equation}
% \label{eq:GSNET}
G = softmax\left( {\frac{{{Q_S}K_E^T}}{{\sqrt {{d_k}} }} + B} \right){V_E},
\end{equation}
where \({d_k}\) indicates the dimension of queries and keys, and \emph{B} is a relative position bias~\cite{liu2021swin}.

\begin{figure*}[ht]
% \Centering
  \includegraphics[width=\textwidth]{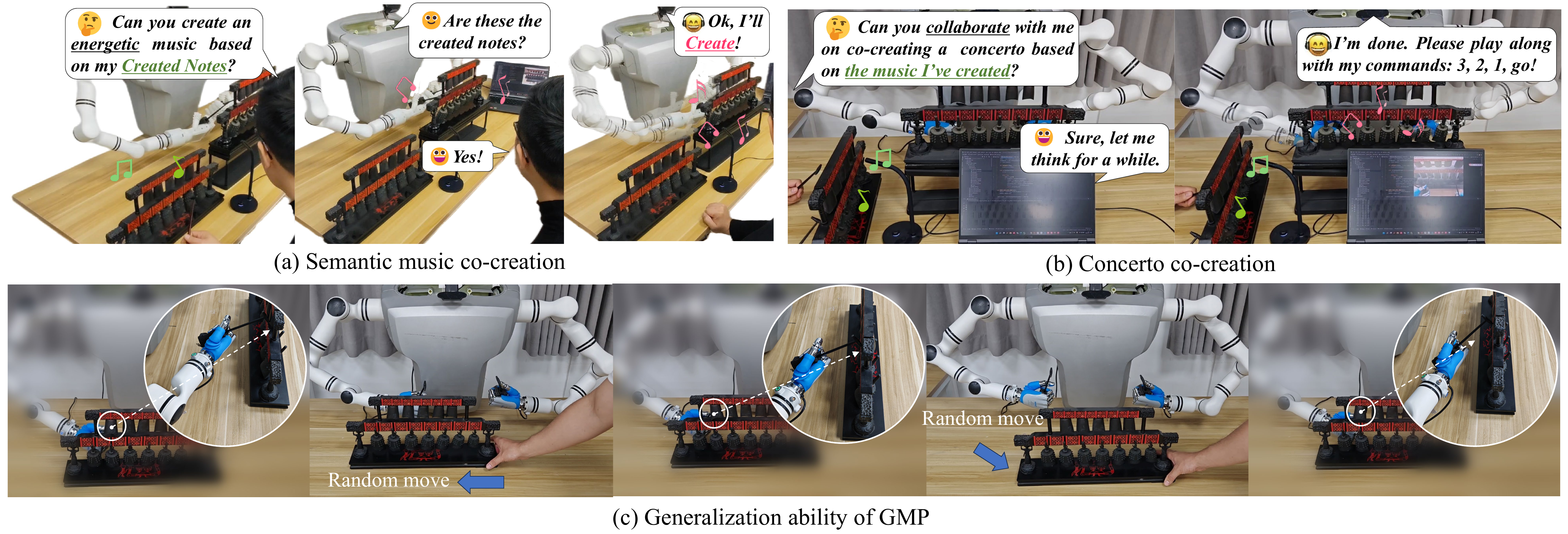}
  
    % \captionsetup{justification=justified}
            \vspace{-0.2cm}
  \caption{\textbf{Experiment in real-world scenarios.} (a) and (b): the ability of co-creation for semantic music and concertos; (c): we randomly adjust the position of the chime, and the robot successfully strikes the designated bell without disrupting its sound.}
\label{fig:real}
\vspace{-0.4cm}
\end{figure*}

In addition to exploiting local and global features, we apply a channel-wise attention module, namely AWCA~\cite{li2020adaptive}, to reallocate channel-wise feature responses by integrating correlations between channels. To improve feature compatibility between Streams 1 and 2, we apply the PSNL~\cite{li2020adaptive} module (Figure~\ref{fig:GSA}), a convolution-based attention mechanism that captures long-range dependencies, analogous to the shift window scheme of Swin Transformer. GSA supplies visual features to the action head:
\begin{equation}
\label{eq:GSNET}
    h_t = \mathrm{GSA}(\mathcal{O}_t)
    = \mathrm{Pool}\!\left(Con[G \odot P_s(A_w(F_1)); Avg(F_1)]\right),
\end{equation}
where \(h_t\) is the visual feature supplied to the mixture-density action head, \(Con[\cdot ;\cdot ]\) indicates channel-wise concatenation, \(\odot\) represents Hadamard product, \(Avg(\cdot)\) is average pooling, and \(A_w\) and \(P_s\) denote AWCA and PSNL, respectively. The action distribution is then predicted by:
\begin{equation}
\label{eq:mixhead}
\{\pi_k,\mu_k,\Sigma_k\}_{k=1}^{K}=f_\theta(h_t,N_t^r,c_t).
\end{equation}
Training uses the mixture negative log-likelihood and auxiliary stabilization loss in Eq.~\ref{eq:loss}; the encoder is optimized jointly with the mixture-density action head.

\paragraph{Conditional mixture-density action generation.}
In robotic musical performance, different trajectories can often realize the same target note: the robot may approach from the left, strike from above, or use a lighter contact depending on the visual layout and recent motion. A deterministic MSE policy tends to average these distinct modes and produce physically ambiguous actions. GMP instead directly predicts a conditional mixture distribution:

\begin{table}[!t]
  \centering
  \vspace{-2pt}
  \scriptsize
  \setlength{\tabcolsep}{1.8pt}
  \renewcommand{\arraystretch}{0.92}
  \resizebox{0.98\columnwidth}{!}{%
  \begin{tabular}{l|cccc|cccc}%
\hline\thickhline
\rowcolor{mygray}
& \multicolumn{4}{c|}{Semantic} & \multicolumn{4}{c}{Concerto}  \\
      \rowcolor{mygray}
\multirow{-2}{*}{Methods} & Int. & Nov. & Coh. & AVG $\uparrow$ & Comp. & Nov. & Coh. & AVG $\uparrow$ \\
\hline\hline
\multicolumn{9}{l}{\textit{ManiSkill2-1st}} \\
\hline
Qwen-vl  & 49.1 & 45.3 & 56.1 & 50.2 & 38.0 & 42.5 & 41.1 & 40.5 \\
F-Qwen   & 59.5 & 65.1 & 66.0 & 63.5 & 48.5 & 50.1 & 51.3 & 50.0 \\
\hline\hline
\multicolumn{9}{l}{\textit{Diffusion Policy}} \\
\hline
Qwen-vl  & 51.5 & 57.3 & 55.0 & 54.6 & 32.5 & 51.1 & 42.0 & 41.9 \\
F-Qwen   & 59.3 & 64.1 & 61.5 & 61.6 & 42.1 & 56.3 & 52.5 & 50.3 \\
\hline\hline
\multicolumn{9}{l}{\textit{\(\pi_{0.5}\)}} \\
\hline
Qwen-vl  & 56.5 & 55.5 & 63.1 & 58.4 & 51.5 & 47.1 & 57.5 & 52.0 \\
F-Qwen   & 61.5 & 62.1 & 53.5 & 59.0 & 41.5 & 53.1 & 43.2 & 45.9 \\
\hline\hline
\multicolumn{9}{l}{\textit{GR00T}} \\
\hline
Qwen-vl  & 56.1 & 51.5 & 47.1 & 51.6 & 38.3 & 47.5 & 38.5 & 41.4 \\
F-Qwen   & 63.3 & 65.1 & 70.5 & 66.3 & 54.3 & 52.0 & 61.1 & 55.8 \\
\hline\hline
\multicolumn{9}{l}{\textit{\textbf{our GMP}}} \\
\hline
Qwen-vl  & 66.1 & 61.5 & 57.1 & 61.6 & 48.3 & 57.5 & 48.5 & 51.4 \\
\textbf{F-Qwen} & \textbf{75.1} & \textbf{71.0} & \textbf{78.3} & \textbf{74.8} & \textbf{58.3} & \textbf{65.5} & \textbf{62.3} & \textbf{62.0} \\
% \hline\thickhline
  \end{tabular}%
  }%
  \vspace{-0.2cm}
\caption{\textbf{Co-creation performance comparison.} F-Qwen denotes the fine-tuned Qwen-vl with semantic anchoring. Intent, Novelty, Coher., and Compl. report intent alignment, creative contribution/novelty, musical coherence, and human-robot complementarity, respectively. AVG is the average score over all retained expert ratings, with no trimming of high-variance artistic responses.
  % $Acc_a$ and $Acc_s$ represent the action success rate and tuning accuracy, respectively. Freq denotes the frequency at which the robot executes actions after receiving a command.
  }
  \label{tb:real}
      \vspace{-0.4cm}
\end{table}

\begin{figure}[htbp]
  % \Centering
  % \vspace{-0.4cm}
    \includegraphics[width=\columnwidth]{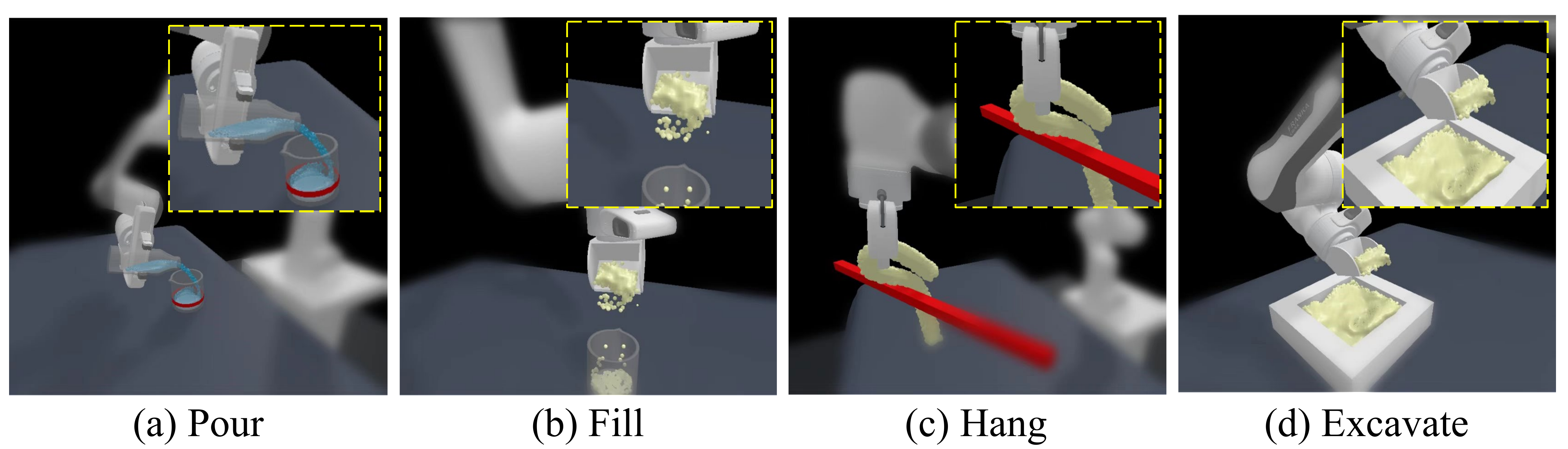}
    
    % \captionsetup{justification=justified, singlelinecheck=false}
    \vspace{-0.3cm}
    \caption{\textbf{Four manipulation tasks in ManiSkill2.} We validated the effectiveness of GMP in each of the four tasks on the virtual platform ManiSkill2. }
  \label{maniskill}
  \vspace{-0.2cm}
  \end{figure}

  \begin{figure}[!t]
    \centering
    % \vspace{-0.4cm}
      \includegraphics[width=0.8\columnwidth]{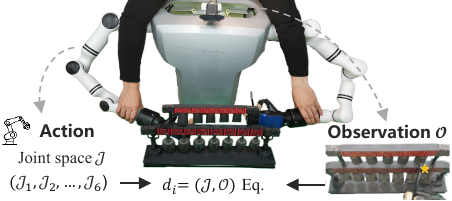}
                  \put(-80,7.0){\makebox(0,0)[l]{\fontsize{7}{8}\selectfont\ref{eq:collection}}}
      % \captionsetup{justification=justified, singlelinecheck=false}
        \vspace{-0.2cm}
      \caption{\textbf{Human demonstration collecting.} For every RGB frame the corresponding action shows how the hand moves from its starting place to the goal spot with the chosen end-effector pose.}
    \label{fig:collection}
    \vspace{-0.5cm}
    \end{figure}

For an action segment \(\tau_t=[a_t,\ldots,a_{t+H}]\in\mathbb{R}^{H\times6}\), GMP predicts:
\begin{equation}
\label{eq:fitting}
p_\theta(\tau_t \mid \mathcal{O}_t,N^r_t,c_t)
=
\sum_{k=1}^{K}\pi_k
\mathcal{N}\!\left(\tau_t;\mu_k,\Sigma_k\right),
\end{equation}  
where \(\pi_k\), \(\mu_k\), and \(\Sigma_k\) are outputs of the neural policy and depend on the current observation and target note plan. Each component represents a latent \textbf{action mode} rather than a joint. The full covariance $\Sigma_k$ captures inter-joint coupling within that mode. We use diagonal or low-rank covariance in implementation for numerical stability and real-time inference, while retaining a multimodal distribution over action vectors.

The policy is trained by negative log likelihood:
\begin{equation}
\mathcal{L}_{\mathrm{MDN}}
=
-\log
\sum_{k=1}^{K}
\pi_k
\mathcal{N}\!\left(\tau^{\mathrm{gt}};\mu_k,\Sigma_k\right),
\end{equation}
with an auxiliary stabilization term
\begin{equation}
\label{eq:loss}
\mathcal{L}
=
\mathcal{L}_{\mathrm{MDN}}
+
\lambda
\left\|
\sum_{k=1}^{K}\pi_k\mu_k-\tau^{\mathrm{gt}}
\right\|_2^2 .
\end{equation}
At inference time, the trained policy computes the mixture parameters once and selects either the most likely mode or the mixture expectation:
\begin{equation}
\label{eq:regression}
\tau^*
=
\mu_{k^*},
\quad
k^*=\arg\max_k \pi_k
\quad
\text{or}
\quad
\tau^*=\sum_{k=1}^{K}\pi_k\mu_k .
\end{equation}
Online computation is therefore one GSA pass plus one mixture-head evaluation; multimodality is represented directly in the output distribution.

\subsection{Experimental Evaluation}
\label{sec:setup}
\subsubsection{Hardware Setup and Simulation Platform} 
\label{subsec:hardware}
The hardware platform for real-world scenarios is presented in the form of a desktop robot. In particular, to mimic the muscle tissue of human fingers and the bending structure of joints, we designed a flexible dexterous hand, thereby making the robotic chime-striking sound more consistent with that of a human hand striking the chime. To verify the generalization of our method, we tested the algorithm in real-life scenarios and on the ManiSkill2 simulation platform (Figure~\ref{fig:real} and ~\ref{maniskill}).

\begin{figure*}[h]
\centering
\includegraphics[width=\textwidth]{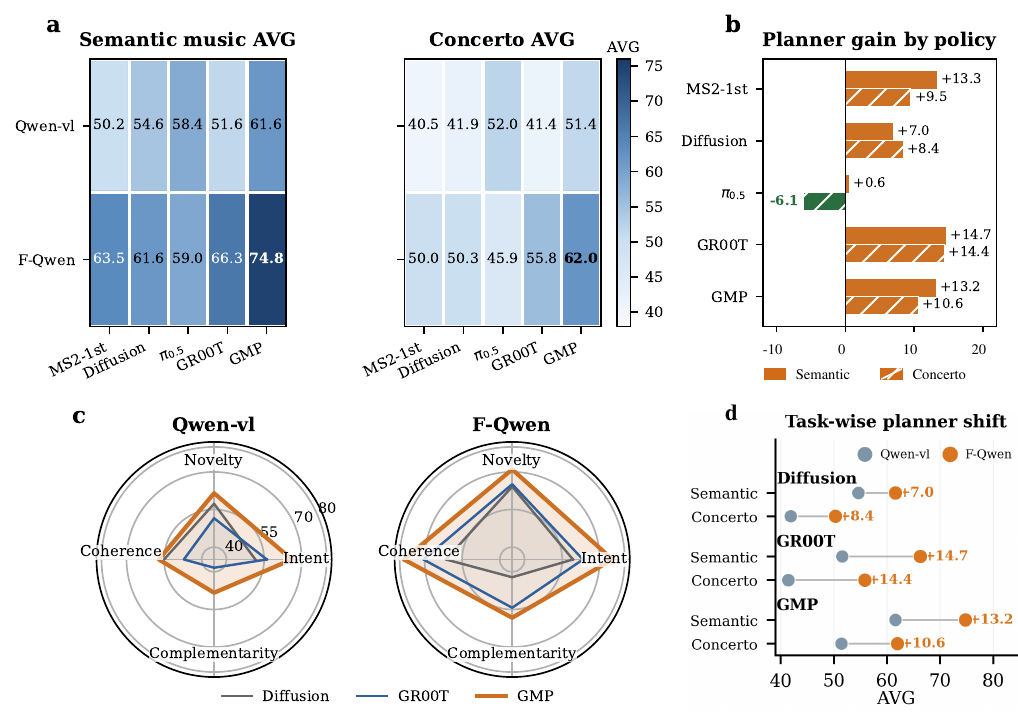}
\vspace{-1.0cm}
\caption{\textbf{Co-creation evaluation across policies and planners.}
\textbf{a}, Policy-planner heatmaps for retained AVG scores in semantic music co-creation and concerto co-creation. 
\textbf{b}, Planner-side gain, computed as F-Qwen minus Qwen-vl under the same action policy; orange indicates improvement and green indicates decrease. MS2-1st denotes ManiSkill2-1st.
\textbf{c}, Descriptive radar profiles comparing Qwen-vl and F-Qwen for Diffusion Policy, GR00T, and GMP across intent alignment, novelty, coherence, and complementarity.
\textbf{d}, Task-wise planner shift for representative policies. Each row reports one co-creation setting under a fixed action policy; left and right endpoints denote Qwen-vl and F-Qwen retained AVG scores, and endpoint labels indicate the planner-side gain.}
\label{fig:cocreation_eval}
\vspace{-0.3cm}
\end{figure*}

\begin{figure}[htbp]
  \centering
  \includegraphics[width=\columnwidth]{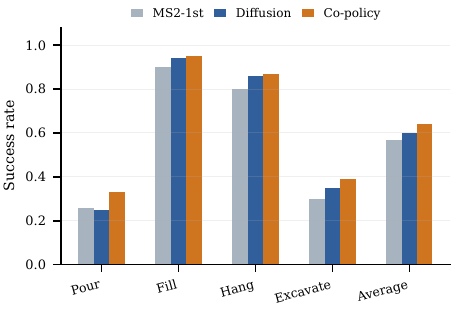}
  % \vspace{-0.8cm}
  \caption{\textbf{Simulation generalization across ManiSkill2 tasks.} Success rates on Pour, Fill, Hang, Excavate, and their average. MS2-1st denotes the first-place ManiSkill2-style policy used as the simulation reference.}
  \label{fig:mani_bar}
  \vspace{-0.4cm}
\end{figure}

\subsubsection{Dataset and Evaluation Criteria}
\label{subsec:data}
To validate the effectiveness of Co-policy, 350 real-world trajectories were collected for training. Each trajectory includes an execution path and an RGB image $\mathcal{O}$ (captured pre-action) linked to a target action:
\begin{equation}
\label{eq:collection}
d_{i}=(\mathcal{J,O}), 
\end{equation}
where $\mathcal{J}$ is the robot-arm joint space (Figure~\ref{fig:collection}). Real-world evaluation uses two criteria. \textit{Co-creation}: 10 professional chime players scored the generated clips along four aggregate axes--intent alignment, creative contribution/novelty, musical coherence, and response fluency/complementarity--with a maximum score of 100 (Table~\ref{tb:real}). These axes operationalize the core co-creation requirements of intent alignment, creative contribution, musical coherence, and physically responsive performance while retaining all expert ratings.

\paragraph{Scoring protocol and statistics.}
Following reviewer guidance, we \textbf{do not discard any extreme scores}: in artistic evaluation an outlier may signal a genuinely \enquote{out-of-distribution} improvisation rather than noise, so every rating is retained. The listening study is blinded and randomized: each expert evaluates anonymized clips from all methods without access to method identities, and each condition contains the same number of clips. We report the retained aggregate scores in Table~\ref{tb:real} and summarize the policy-planner trends in Figure~\ref{fig:cocreation_eval}.

\noindent\textit{Generalization}: action success rate \(Acc_a\) (accurate strikes without vibration/sound interference) and tuning accuracy \(Acc_t\) (computed by the Methods note-recognition pipeline):
\begin{equation}
\label{eq:acc}
 Acc= Acc_a \times Acc_t,\quad
 Acc_t=\operatorname{clip}_{[0,1]}\!\left(1-\frac{|F_s-F_t|}{|F_s-F_n|}\right),
\end{equation}
where $F_s$ is the striking frequency, $F_t$ is the target frequency, and $F_n$ is the adjacent note frequency immediately before or after the striking note; the clipping keeps tuning accuracy within \([0,1]\). The real-robot objective metrics are action success, tuning accuracy, combined accuracy, and post-command action frequency (Table~\ref{tb:chime}). For the simulation platform, we utilize ManiSkill2~\cite{gu2023maniskill2} only as a generalization sanity check for the visuomotor policy across four soft-body manipulation tasks.

% \begin{table}[!t]
% \vspace{0.1cm}
% \begin{center}
% \small
% \renewcommand\tabcolsep{3.0pt}
% \begin{tabular}{c|cccc|c}
% \rowcolor{mygray}
% \toprule
%  \textbf{Methods}& \textbf{Pour}$\uparrow$   & \textbf{Fill} $\uparrow$ & \textbf{Hang} $\uparrow$ & \textbf{Exca} $\uparrow$ & \textbf{Average} $\uparrow$    \\  
%  \midrule
% {ManiSkill2-1st} &  0.26 & 0.90 &0.80 & 0.30 & 0.57\\  
                                  
% {Diffusion Policy} &  0.25 & 0.94 &0.86 & 0.35 & 0.60 \\

% \rowcolor[HTML]{D7F6FF}
% \textbf{GMP(ours)}  &  \textbf{0.33} & \textbf{0.95} &\textbf{0.87} & \textbf{0.39} & \textbf{0.64} \\

% \bottomrule
% \end{tabular}
% \vspace{-0.3cm}
% \caption{ \textbf{Evaluation results of simulation scenarios} on the method of the 1st in ManiSkill2 challenge, diffusion policy and our GMP. Exca denotes the Excavate task.}
% \vspace{-0.6cm}
% \label{mani}
% \end{center}
% \end{table}

% \begin{figure}[t]
% % \Centering
% % \vspace{-0.4cm}
%   \includegraphics[width=\columnwidth]{co-creation__experiment.pdf}
  
%   % \captionsetup{justification=justified, singlelinecheck=false}
%     \vspace{-0.2cm}
%   \caption{\textbf{Experiment in real-world scenarios}.  we randomly adjust the position of the chime, and the robot successfully strikes the designated bell without disrupting its sound production.}
% \label{fig:real-world}
% \vspace{-0.4cm}
% \end{figure}

\subsubsection{Performance Comparison and Ablation Study}
\label{subsec:performance}
\paragraph{Baselines and configurations.}
For a fair comparison, all baselines consume the same egocentric RGB observation, use the same 350 demonstration trajectories, predict the same six-DoF joint command, and are evaluated under the same action horizon. \textit{ManiSkill2-1st}~\cite{gao2023two} is the first-place soft-body manipulation entry re-trained on our chime demonstrations. \textit{Diffusion Policy}~\cite{chi2023diffusion} uses the public CNN-based U-Net backbone with a DDPM schedule and $T{=}50$--$100$ denoising steps at inference. Because RH20T~\cite{fang2023rh20t} and DROID~\cite{khazatsky2024droid} are datasets rather than single algorithms, we instantiate them as behavior-cloning controls rather than claiming to reproduce those datasets: the RH20T-style BC uses a ResNet visual encoder and MSE action regression, while the DROID-adapted BC uses a transformer-style visual-token encoder and the same MSE action loss. Neither baseline uses additional RH20T or DROID trajectories. In the co-creation study (Table~\ref{tb:real}), each control policy is paired with either the off-the-shelf \textit{Qwen-vl} or our fine-tuned \textit{F-Qwen} planner so that the contribution of semantic anchoring can be isolated.

\paragraph{Latency measurement.}
The response-speed claim is evaluated under identical hardware, batch size one, and the same action horizon. Table~\ref{tb:chime} reports post-command action frequency, which captures policy-side response speed after the robot receives a command. This controlled comparison focuses on the visuomotor execution loop: GMP inference is included because it requires one encoder pass and one mixture-head evaluation, whereas diffusion-policy baselines require repeated denoising steps under the same action horizon. VLM planning is not included in this post-command frequency; robot communication and actuation delay are not separately isolated; and full speech-to-action interaction latency is not claimed by Table~\ref{tb:chime}.

\paragraph{Ablations for reproducibility.}
To avoid overclaiming, we use ablations as component diagnostics rather than a full factorial sweep. Semantic anchoring is assessed through the Qwen-vl/F-Qwen comparisons, while the action-side ablations remove ST, DN, GSA, or the GMP head from the deployed pipeline. These diagnostics are intended to show, in this study, that the reported co-creation quality does not simply arise from the base VLM and that multimodal action prediction contributes beyond a deterministic MSE policy.

To evaluate the performance of GMP, we conducted comparative experiments in real-world settings and simulation platforms, benchmarking against the first-place entry~\cite{gao2023two} in the ManiSkill2 challenge and the SOTA models. As shown in Figure~\ref{fig:cocreation_eval}, in co-creation experiments, Co-policy outperforms the diffusion policy by 8.5\% (74.8 vs. 66.3) and 6.2\% (62.0 vs. 55.8). After applying pre-inference semantic anchoring to Qwen-vl, the model achieves a 7.4\%--13.3\% improvement in semantic music co-creation and a 3.5\%--9.6\% improvement in concerto co-creation. The negative concerto gain for \(\pi_{0.5}\) suggests that semantic anchoring alone cannot compensate for policy-side execution mismatch. For real-robot chime striking (Table~\ref{tb:chime}), we observe a 15\% performance gain and a higher response frequency than the diffusion-policy baseline under the same action-horizon setting. In the ManiSkill2 simulation sanity check (Figure~\ref{fig:mani_bar}), GMP is treated as a visuomotor generalization sanity check rather than the central claim. Furthermore, ablation studies (Table~\ref{tb:abla}) reveal the contributions of key GMP components: the Swin Transformer stage (32\%), DenseNet block in the encoder (27\%), GSA (21\%), and mixture-density action head (16\%). The main evidence is therefore the real-robot co-creation and chime-striking evaluation.

\begin{figure*}[!t]
  \centering
  \includegraphics[width=0.98\textwidth]{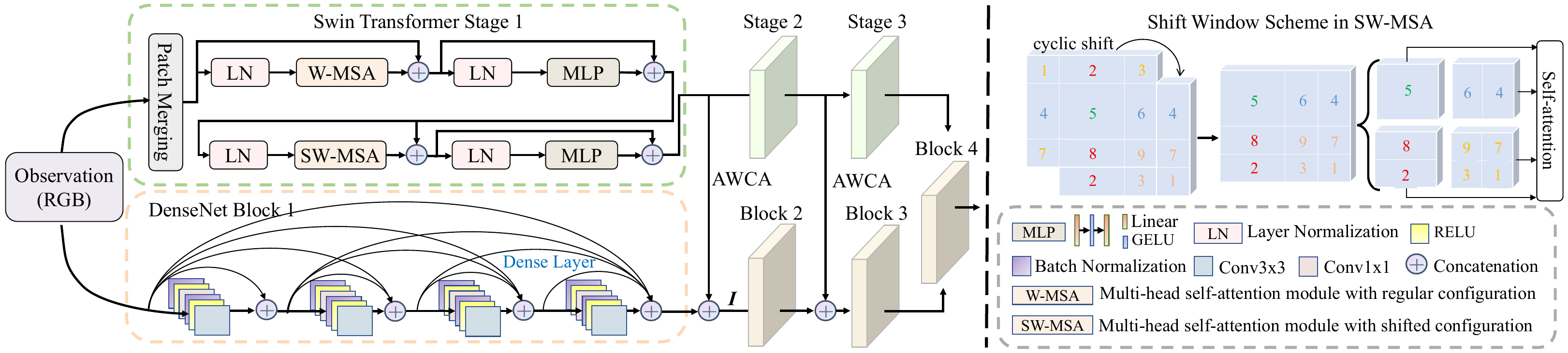}
  \vspace{-0.4cm}
  \caption{\textbf{Encoder structure for visuomotor grounding.} This encoder diagram clarifies how global Swin-Transformer features and local DenseNet features are coupled before guided self-attention. AWCA reweights channel responses and PSNL improves long-range compatibility in the local stream, allowing the GMP head to condition action-mode prediction on both instrument-level layout and contact-level visual evidence.}
  \label{fig:discussion_encoder}
  \vspace{-0.4cm}
\end{figure*}

\section{Discussion}
\label{sec:conclution}
This work reframes robotic musical performance as an embodied co-creation problem rather than a playback problem. The robot is not evaluated only by whether it strikes a requested bell, but by whether it can transform an incomplete human creative seed into a complementary musical response that is semantically aligned, musically coherent, physically executable, and responsive at interaction speed. This distinction is central for embodied AI: musical meaning is generated through the coupling of symbolic intent, acoustic context, visual grounding, and contact-rich action.

The results suggest that physically grounded action generation is a necessary component of human-AI co-creation. A purely symbolic music model can generate plausible notes, but it does not decide whether a physical robot can reach the instrument, avoid occlusion, satisfy timing constraints, or produce the intended sound after contact. Conversely, a low-level controller can strike bells accurately but cannot determine whether the response contributes musically to the human seed. Co-policy therefore separates semantic grounding, constrained musical variation, and visuomotor execution while preserving information flow between them. This separation makes the system interpretable: semantic failures, perceptual localization failures, and execution failures can be inspected independently rather than being collapsed into a single end-to-end error.

\begin{table}[!t]\small
  \centering
  {
  \resizebox{0.96\columnwidth}{!}{
  \setlength\tabcolsep{3pt}
  \renewcommand\arraystretch{1.05}
  \begin{tabular}{c||ccc|c}
  \hline\thickhline
  \rowcolor{mygray}
  Methods& $Acc_a$ $\uparrow$  & $Acc_t$ $\uparrow$ & Acc $\uparrow$ & Freq (Hz) $\uparrow$   \\ \hline\hline
  {ManiSkill2-1st} & 0.60 & 0.73 & 0.44 & 15.2\\  
  {BC (RH20T-style)} & 0.65 & 0.78 & 0.51 & 10.1 \\                                 
  {Diffusion Policy} & 0.68 & 0.83 & 0.56 & 1.01 \\
  {DROID-adapted BC} & 0.62 & 0.75 & 0.47 & 14.2 \\
  \hline
  \textbf{Co-policy (ours)}  &   \textbf{0.78} &  \textbf{0.88}  & \textbf{0.69} &  {\textbf{18.6}}
  \end{tabular}
  }
  }
  \vspace{-8pt}
  \caption{\textbf{Evaluation of chime striking.} \(Acc_a\), \(Acc_t\), Acc, and Freq denote action success, tuning accuracy, combined accuracy, and post-command frequency.}
  \label{tb:chime}
  \vspace{-0.5cm}
\end{table}

The framework also clarifies the boundary of robotic creativity. The generated response is not a free-form hallucination from the VLM, nor is it a simple lookup from a fixed anchor bank. Instead, the VLM supplies broad semantic generalization over style, intent, and visual context, while the semantic anchors and playability filters define the domain in which variation is musically and physically meaningful. Operationally, the prompt acts as a constraint interface: user intent, seed notes, playable-note limits, timing, and hand-state information are passed through semantic-anchor guidance, and the VLM is required to return a structured response containing notes with beats, robot role, timing, and hand assignment. Creativity in this setting is therefore guided variation: the robot may alter rhythm, accompaniment role, or melodic contour, but it must remain within the constraints imposed by the instrument, the human seed, and the executable action space. This interpretation is conservative but important for embodied AI, where unconstrained novelty can easily become an unplayable or unsafe physical command.

The comparison with diffusion-policy baselines should be understood in the same embodied context. Co-policy does not claim that mixture-density policies universally dominate diffusion models. Recent accelerated diffusion variants can reduce sampling cost, and diffusion models remain powerful for complex trajectory distributions. The contribution of GMP is narrower: for short-horizon chime striking, a conditional mixture-density head preserves multiple feasible latent action modes while requiring only one policy evaluation after the command has been received. This is useful when the interaction loop is constrained by real-time musical timing, where repeated denoising can be expensive and deterministic behavior cloning may collapse to an averaged, weak-contact action.

\begin{table}[!t]\small
  \small
  \centering
  {
  \resizebox{0.98\columnwidth}{!}{
  \setlength\tabcolsep{3pt}
  \renewcommand\arraystretch{1.05}
  \begin{tabular}{cccc|cccc}
  \hline\thickhline
  \rowcolor{mygray}
   ST & DN & GSA & GMP head   & $Acc_a$ $\uparrow$ & $Acc_t$ $\uparrow$ & Acc $\uparrow$ & $\Delta$Acc $\downarrow$ \\ \hline\hline
   - &  \checkmark & \checkmark & \checkmark & 0.57 &0.65  & 0.37 & 0.32 \\
   \checkmark &  - & \checkmark & \checkmark & 0.62 &0.68  & 0.42 & 0.27 \\
   \checkmark &  \checkmark & - & \checkmark & 0.68 &0.70  & 0.48 & 0.21 \\
   \checkmark &  \checkmark & \checkmark & - & 0.71 &0.75  & 0.53 & 0.16 \\
  \hline
   \checkmark &  \checkmark & \checkmark & \checkmark  & \textbf{0.78} &  \textbf{0.88}  & \textbf{0.69} & \textbf{0.00} \\
  \end{tabular}
  }
  }
  \vspace{-8pt}
  % \captionsetup{font=small}
  \captionof{table}{\textbf{Ablation study of core GMP modules.} We test the Swin Transformer stage (ST), DenseNet block (DN), Guided Self-Attention mechanism (GSA), and GMP mixture-density action head on 30 randomly positioned chime samples. \(\Delta\)Acc is the absolute Acc drop from the full model.}
  \label{tb:abla}
  \vspace{-0.5cm}
\end{table}

\begin{figure*}[htbp]
  \centering
  \begin{subfigure}[t]{0.31\textwidth}
  \centering
  \includegraphics[width=\linewidth]{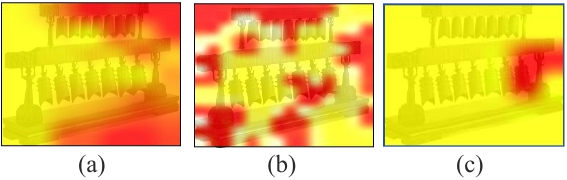}
  \caption{Attention map}
  \end{subfigure}
  \hfill
  \begin{subfigure}[t]{0.31\textwidth}
  \centering
  \includegraphics[width=\linewidth]{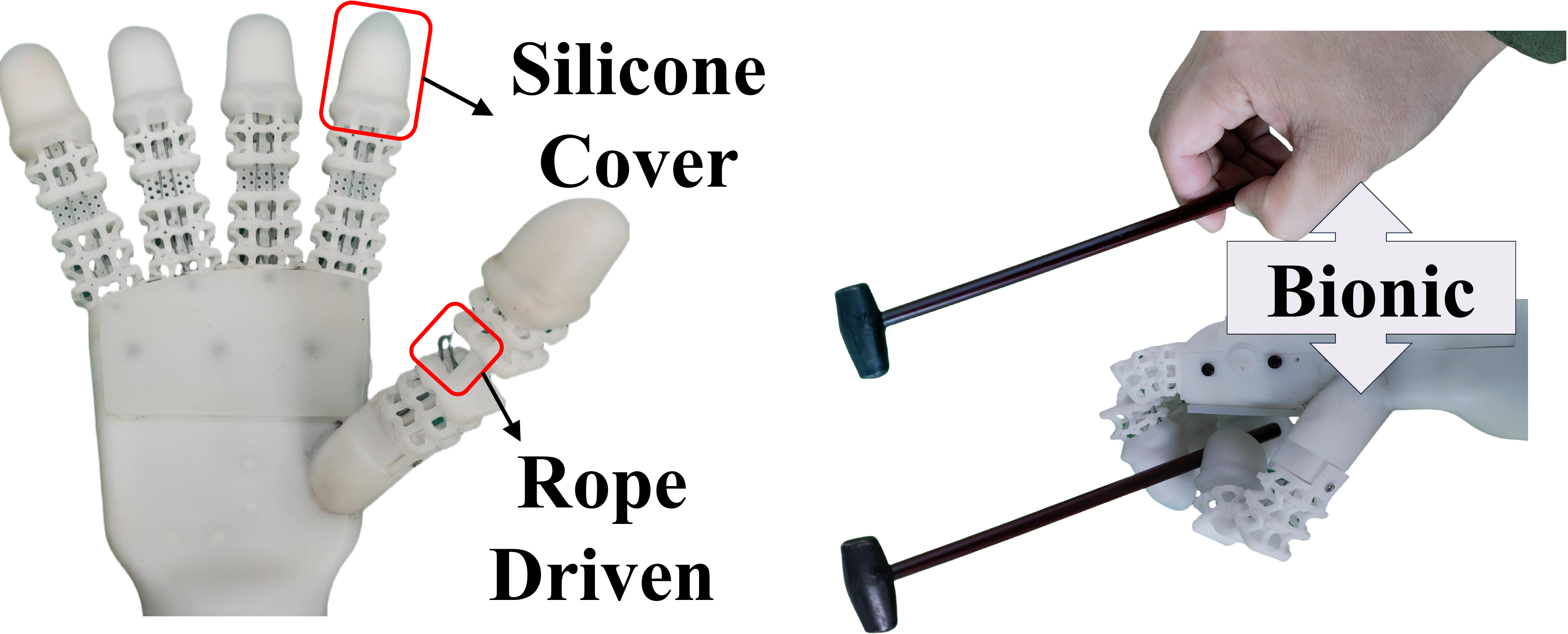}
  \caption{Compliant hand}
  \end{subfigure}
  \hfill
  \begin{subfigure}[t]{0.31\textwidth}
  \centering
  \includegraphics[width=\linewidth]{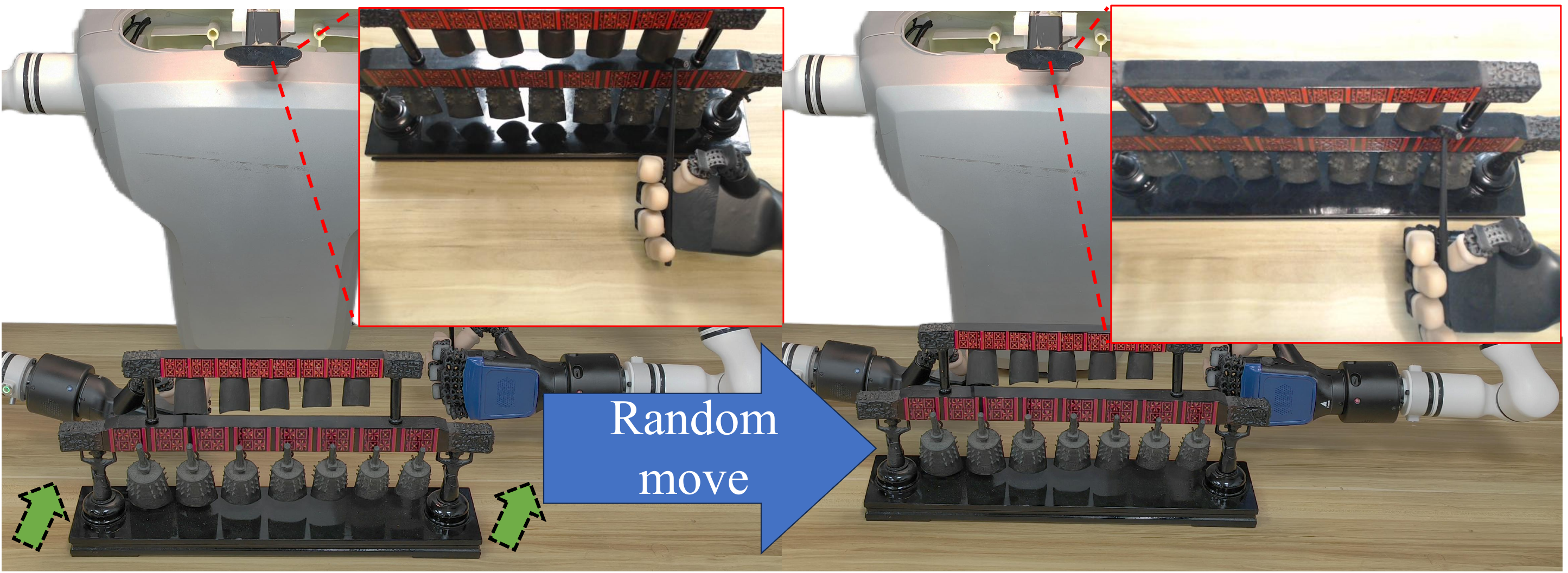}
  \caption{Robot perspective}
  \end{subfigure}
  \caption{\textbf{Discussion-oriented diagnostics.} Attention visualization helps inspect whether the visual encoder grounds actions around target bells and contact-adjacent regions. The compliant hand illustrates the soft-contact design used to approximate human chime striking, while the robot-perspective view shows why small visual differences can still induce multiple feasible action modes. Together, these diagnostics clarify why Co-policy treats creativity, grounding, and execution as coupled but separately inspectable components.}
  \label{fig:discussion_diagnostics}
  \vspace{-0.5cm}
\end{figure*}

This design choice is linked to the encoder diagnostic summarized in Figure~\ref{fig:discussion_encoder}. In a musical robot, local texture around a bell edge and global layout of the chime frame are both relevant: the former affects contact localization, while the latter affects target selection and reachability. The encoder should therefore be read as a practical mechanism for improving feature compatibility before action-mode prediction, not as an independent claim about musical intelligence. Its role is to reduce perceptual ambiguity before the mixture-density policy decides among alternative feasible strikes.

Several limitations remain. First, the current system is validated primarily on chime performance, so the musical space is bounded by the available notes, the anchor bank, and the visual geometry of the instrument. Second, acoustic note extraction can be affected by servo noise, hammer impact, and chime reverberation. The present system treats audio as a high-level symbolic cue for planning rather than as a dense feedback signal for contact control. It does not yet include closed-loop tactile or force sensing after impact, so mis-strikes are detected through pitch and action-success metrics rather than corrected during contact. Third, the evaluation focuses on short interaction windows; long-term improvisation would require memory, adaptation to a specific human partner, and higher-level musical form planning. These limits are also reflected in the diagnostic views in Figure~\ref{fig:discussion_diagnostics}: attention must remain concentrated near playable contact regions, the hand design can only approximate human compliance, and small visual changes from the robot perspective can still correspond to different feasible striking trajectories.

These limitations point to clear future directions. Source-separated acoustic perception could distinguish human notes, robot impacts, and reverberant tails more reliably. Contact sensing and force feedback could enable online recovery from weak strikes or unintended collisions. Larger anchor banks and preference models could support more diverse musical styles while preserving playability constraints. Finally, multi-turn co-creation with memory would allow the robot to develop motifs, respond to repeated human themes, and participate in ensemble-level structure rather than isolated call-and-response phrases.

\section{Methods}
\subsection{Problem Formulation and Overview}
At interaction step \(t\), Co-policy receives a human instruction \(\mathcal{I}_t\), symbolic seed notes \(N^u_t\), and an egocentric RGB observation \(\mathcal{O}_t\), from which it estimates physical constraints \(c_t\). The system first plans a robot musical response \(N^r_t\) and then predicts an executable action segment \(\tau_t=[a_t,\ldots,a_{t+H}]\):
\begin{equation}
\label{eq:copolicy_factorization}
N^r_t=f_{\mathrm{plan}}(\mathcal{I}_t,N^u_t,\mathcal{O}_t,c_t;\mathcal{A}),\quad
\tau_t=f_\theta(\mathcal{O}_t,N^r_t,c_t),
\end{equation}
where \(\mathcal{A}\) is the semantic anchor bank and \(f_\theta\) is the Gaussian-Mixture Visuomotor Policy. This factorization allows the semantic response, musical validity, and physical execution to be evaluated separately.

\subsection{Semantic Grounding and Constrained Musical Planning}
Each anchor \(a_j\in\mathcal{A}\) contains a musical seed, score image, style descriptor, tempo or meter cue, playability tag, and expected structured plan. Given the query \(q_t=(\mathcal{I}_t,N^u_t,\mathcal{O}_t)\), the planner retrieves:
\begin{equation}
\label{eq:anchor_retrieval}
\mathcal{A}_t=\mathrm{TopM}_{a_j\in\mathcal{A}}
\cos\!\left(\phi(q_t),\phi(a_j)\right),
\end{equation}
where \(\phi(\cdot)\) denotes the multimodal embedding. The retrieved anchors condition Qwen-vl, which must output a structured plan rather than free-form prose. Candidate note responses are selected by a constrained objective:
\begin{equation}
\label{eq:planner_objective}
\begin{aligned}
N^r_t=\arg\max_{N\in\mathcal{C}(N^u_t,c_t)}
&\ \alpha S_{\mathrm{sem}}(N)
+\beta S_{\mathrm{mus}}(N,N^u_t)\\
&+\gamma S_{\mathrm{nov}}(N,N^u_t)
-\eta C_{\mathrm{phy}}(N,c_t),
\end{aligned}
\end{equation}
where \(\mathcal{C}\) denotes reachable and timing-valid candidates, \(S_{\mathrm{sem}}\) measures intent alignment, \(S_{\mathrm{mus}}\) musical coherence, \(S_{\mathrm{nov}}\) creative deviation from the seed, and \(C_{\mathrm{phy}}\) physical cost. Live audio is converted to \(N^u_t\) through onset detection, pitch tracking, and rhythm quantization; this audio processing is distinct from the post-command policy frequency reported in Table~\ref{tb:chime}.

\subsection{Gaussian-Mixture Visuomotor Policy}
The visual feature is computed by coupling global scene context and local contact evidence:
\begin{equation}
\label{eq:gsa_feature}
h_t=\mathrm{GSA}\!\left(g_{\mathrm{Swin}}(\mathcal{O}_t),
g_{\mathrm{Dense}}(\mathcal{O}_t)\right),
\end{equation}
where AWCA reweights channels and PSNL improves long-range feature compatibility before guided attention. GMP then predicts a conditional mixture over complete short-horizon action segments:
\begin{equation}
\label{eq:gmp_method}
p_\theta(\tau_t\mid h_t,N^r_t,c_t)=
\sum_{k=1}^{K}\pi_k(h_t,N^r_t,c_t)
\mathcal{N}\!\left(\tau_t;\mu_k,\Sigma_k\right).
\end{equation}
Each component is a latent action mode, not a robot joint. \(K=6\) is used as a practical capacity hyperparameter, and \(\Sigma_k\) is constrained to positive diagonal or low-rank form for stable real-time inference.

\subsection{Training, Evaluation, and Diagnostics}
Demonstrations are tuples \(d_i=(\mathcal{O}_i,N^r_i,c_i,\tau_i^{\mathrm{gt}})\). The policy is trained by mixture negative log likelihood plus stabilization:
\begin{equation}
\label{eq:method_loss}
\begin{aligned}
\mathcal{L}_{\mathrm{GMP}}
=&-\sum_i\log\sum_{k=1}^{K}\pi_{ik}
\mathcal{N}\!\left(\tau_i^{\mathrm{gt}};\mu_{ik},\Sigma_{ik}\right)\\
&+\lambda\sum_i\left\|
\sum_{k=1}^{K}\pi_{ik}\mu_{ik}
-\tau_i^{\mathrm{gt}}
\right\|_2^2 .
\end{aligned}
\end{equation}
At inference, the executable segment is \(\tau_t^*=\mu_{k^*}\), \(k^*=\arg\max_k\pi_k\), or the mixture expectation \(\sum_k\pi_k\mu_k\). Post-command frequency measures only this policy-side loop after an executable command has been received.

The real-robot platform uses an egocentric camera, microphone, two six-degree-of-freedom arms, and compliant dexterous hands. We collect 350 real-world demonstrations and adapt all baselines to the same observation space, action horizon, and six-DoF command interface. Co-creation is evaluated by blinded expert ratings on intent alignment, creative contribution, musical coherence, and complementarity. Objective striking metrics are:
\begin{equation}
\label{eq:method_metrics}
\begin{aligned}
Acc&=Acc_a Acc_t,\\
Acc_t&=\operatorname{clip}_{[0,1]}\!\left(1-\frac{|F_s-F_t|}{|F_s-F_n|}\right),\\
e_t&\in\{E_{\mathrm{sem}},E_{\mathrm{vis}},E_{\mathrm{exec}}\},
\end{aligned}
\end{equation}
where \(F_s\), \(F_t\), and \(F_n\) are the struck, target, and adjacent-note frequencies, and \(Acc_t\) is clipped to \([0,1]\). The error label \(e_t\) separates semantic-planning, visual-localization, and execution-contact failures. ManiSkill2 results are used only as a secondary visuomotor generalization check.
\section{Data Availability}
The project webpage, source code, demonstration videos, and supplementary materials are publicly available at \textcolor{magenta}{\url{https://xtli12.github.io/Co-policy/docs/}}. Processed robot demonstrations, generated note plans, anonymized expert ratings, evaluation scripts, prompt templates, semantic-anchor schema, and trained-model configuration files are provided through the public project repository where release permissions allow. Raw videos that may contain identifiable participants are shared only in anonymized, cropped, or consent-permitted form. The semantic-anchor schema and JSON output format are described in the Methods to support reproduction independent of the full raw video release.
\section{Code Availability}
The source code, webpage implementation, prompt templates, and configuration files associated with Co-policy are publicly available through the project repository linked from \textcolor{magenta}{\url{https://xtli12.github.io/Co-policy/docs/}}.
\section{Acknowledgements}
The authors thank the participating musicians and annotators for assisting with real-robot evaluation and perceptual scoring.
\section{Author Contributions}
X.L. designed and implemented Co-policy, conducted experiments, and drafted the manuscript. W.H. and M.Y. contributed to system design, experimental analysis, and manuscript revision. Z.L. and J.Xie contributed to robotic platform construction and evaluation. J.Xuan and M.L. supervised the project, provided guidance, and revised the manuscript. All authors reviewed and approved the manuscript.
\section{Competing Interests}
The authors declare no competing interests.
% \clearpage
\bibliography{reference}

\end{document}